\documentclass{article}
\usepackage{spconf,amsmath,graphicx}
\usepackage[hyphens]{url}
\usepackage{comment}
\usepackage{caption, subfig}
\title{PROFICIENCY ASSESSMENT OF L2 SPOKEN ENGLISH USING WAV2VEC 2.0}
\name{Stefano Bannò$^{1,2}$, Marco Matassoni$^{1}$}
\address{$^{1}$Fondazione Bruno Kessler, Trento, Italy\\
$^{2}$University of Trento, Trento, Italy}
\copyrightnotice{Accepted at SLT 2022. Copyright notice: 978-1-6654-7189-3/22/\$31.00~\copyright2023 IEEE}
\begin{document}
%
\maketitle
\begin{abstract}
The increasing demand for learning English as a second language has led to a growing interest in methods for automatically assessing spoken language proficiency. Most approaches use hand-crafted features, but their efficacy relies on their particular underlying assumptions and they risk discarding potentially salient information about proficiency. Other approaches rely on transcriptions produced by ASR systems which may not provide a faithful rendition of a learner's utterance in specific scenarios (e.g., non-native children's spontaneous speech). Furthermore, transcriptions do not yield any information about relevant aspects such as intonation, rhythm or prosody. In this paper, we investigate the use of wav2vec 2.0 for assessing overall and individual aspects of proficiency on two small datasets, one of which is publicly available. We find that this approach significantly outperforms the BERT-based baseline system trained on ASR and manual transcriptions used for comparison.
\end{abstract}
\section{Introduction}
\label{sec:intro}

With the increasing number of learners of English as a second language (L2) worldwide, there has been a growing demand for automated spoken language assessment systems both for formal settings, such as internationally recognised language tests, and for practice situations in Computer Assisted Language Learning (CALL). In fact, compared to human graders, automatic graders can ensure greater consistency and speed at a lower cost, since the recruitment and training of new human experts are expensive and can offer only a small increase in performance~\cite{zhang2013}. \par
In automatic assessment of L2 spoken language proficiency, input sequential data from a learner is used to predict a grade or a level with respect to holistic proficiency or specific aspects of proficiency~\cite{iwashita2008, dejong2012}. The input may consist of acoustic features, recognised words, phones and/or time-aligned information, or other information, such as fundamental frequency, 
extracted directly from the audio or from automatic speech recognition (ASR) transcriptions. 
Most approaches in the literature extract sets of hand-crafted features related to specific aspects of proficiency, such as fluency~\cite{strik1999automatic}, pronunciation \cite{chen}, prosody~\cite{coutinho2016assessing} and text complexity~\cite{bhat2015automatic}, which are then fed into graders to predict analytic scores targeting those specific aspects. A similar approach can be used by concatenating multiple hand-crafted features targeting more than one aspect in order to produce overall feature sets, which are then passed through graders to predict holistic grades, as shown in~\cite{muller2009automatically, crossley2013applications, wang2018towards, liu2020dolphin}. The effectiveness of hand-crafted features for grading either single aspects or overall proficiency heavily relies on their particular underlying assumptions, and they risk discarding potentially salient information about proficiency. This issue for holistic grading has been addressed by replacing hand-crafted features with automatically derived features for holistic grading prediction, either through an end-to-end system~\cite{chen2018end} or in multiple stages~\cite{takai2020deep, cheng2020asr}. Other studies have employed graders that are trained on holistic grades but are defined with both their inputs and topology adapted to focus on specific aspects of proficiency, such as pronunciation~\cite{kyriakopoulos2018deep}, rhythm~\cite{kyriakopoulos2019deep} and text~\cite{wang2021, raina2020universal}. In these cases, an evident limitation is the lack of information concerning aspects of proficiency that are not included in the input data fed to the grader, although it is possible to combine multiple graders targeting different aspects, as shown in \cite{banno2022view}. This is particularly true for systems using ASR transcriptions for two reasons: first, they contain a certain word error rate (WER) and may not faithfully render the contents of a learner's performance; secondly, although transcriptions might preserve some information about pronunciation (e.g., in the ASR confidence scores), they do not yield any information about other relevant aspects of a learner's performance, e.g., intonation, rhythm or prosody. Instead, they remain a valuable resource for highly specific tasks in CALL applications, such as spoken grammatical error correction and feedback~\cite{lu2022}. \par
In this work, to address these issues and limitations, we propose an approach based on self-supervised speech representations using wav2vec 2.0~\cite{baevski, hsu}. Recent studies have demonstrated that self-supervised learning (SSL) is an effective approach in various downstream tasks of speech processing applications, such as ASR, emotion recognition, keyword spotting, speaker diarisation and speaker identification~\cite{baevski, yang21c}. In these studies, contextual representations were applied by means of pre-trained models. Specifically, it has been demonstrated that such models are capable of capturing a wide range of speech-related features and linguistic information, such as audio, fluency, suprasegmental pronunciation, and even syntactic and semantic text-based features for L1, L2, read and spontaneous speech~\cite{shah}. In the field of CALL, SSL has been applied to mispronunciation detection and diagnosis~\cite{peng21, wu21, xu21} and automatic pronunciation assessment~\cite{eesung}, but, to the best of our knowledge, it has not been investigated for the assessment of overall spoken proficiency nor of other specific aspects of proficiency, such as formal correctness, communicative effectiveness, lexical richness and complexity, and relevance.

In this work, we first test the effectiveness of wav2vec 2.0 in predicting the holistic proficiency level of L2 English learners' answers included in a publicly available dataset. Subsequently, we do the same on a non-publicly available learner corpus, which also contains annotations related to single aspects of proficiency (i.e., pronunciation, fluency, formal correctness, communicative effectiveness, relevance and lexical richness and complexity) that we try to predict with specific graders. The baseline system used for comparison is a BERT~\cite{devlin2018} model that takes transcriptions as input. We use only manual transcriptions for our experiments on the publicly available dataset, whereas we use both manual and ASR transcriptions for our experiments on the second dataset. In particular, the manual transcriptions also contain hesitations and fragments of words, which serve as proxies for pronunciation and fluency.

Another aspect that should be highlighted is that we conduct our experiments using a small quantity of training data and still manage to achieve promising results on both datasets.

Section 2 describes the data used in our experiments. Section 3 illustrates the model architectures. Section 4 shows the results of our experiments. Finally, in Section 5, we analyse and discuss the results and reflect upon next steps.

\section{Data}
\label{sec:format}

\subsection{ICNALE}

In order to test our approach, we consider the International Corpus Network of Asian Learners of English (ICNALE)~\cite{ishikawa2011}, a publicly available dataset~\footnote{http://language.sakura.ne.jp/icnale/download.html} comprising written and spoken responses of English learners ranging from A2 to B2 of the Common European Framework of Reference (CEFR) for languages~\cite{cefr2001} and partially of native speakers. The L1s of the speakers are not indicated, but they may be inferred from their countries of origin: China, Hong Kong, Indonesia, Japan, South Korea, Pakistan, Philippines, Singapore, Thailand, and Taiwan. The CEFR levels were assigned prior to collecting the data, as the ICNALE team required all the learners to take an L2 vocabulary size test and to present their scores in English proficiency tests such as TOEFL, TOEIC, IELTS, etc. On the basis of these two scores, the learners were classified into proficiency levels. The spoken section of the corpus consists of two parts: one containing monologues and the other containing dialogues. For our experiments, we only considered the monologues, i.e., 4332 answers lasting between 36 and 69 seconds in which learners are required to express their opinion about two statements on the importance of having a part-time job and on smoking in restaurants.
The available metadata includes manual transcriptions of the learners' answers, personal information about learners' education history, and their assigned CEFR levels. We divide the data into a training set of 3898 answers, a development set and a test set with 217 answers each. For the experiments on this dataset, proficiency assessment is treated as a classification task with five classes: A2, B1\_1, B1\_2, B2, and native speakers (see Table \ref{T:data_icn_descr}). To the best of our knowledge, the ICNALE spoken monologues have only been used in \cite{zhou19}, but in this study, the answers to the two statements are considered and evaluated independently, so no comparison is possible. The experiments described in \cite{banno2022}, instead, only include a section of essays and spoken dialogues.

\begin{table}[h!]

\begin{center}
\begin{tabular}{ l | c | c | c | c }

   & \textbf{Train} & \textbf{Dev} & \textbf{Test} & \textbf{Total} \\ 

\hline
\hline
A2 & 299   & 16 & 17 & 332 \\ \hline
B1\_1 & 792  & 44 & 44 & 880 \\ \hline
B1\_2 & 1681   & 94 & 93 & 1868 \\ \hline 
B2 & 586 & 33 & 33 & 652 \\ \hline
native & 540 & 30 & 30 & 600 \\ \hline \hline
Total & 3898 & 217 & 217 & 4332 \\ \hline

\end{tabular}
\end{center}
\caption{Number of answers for each CEFR proficiency level in ICNALE.}
\label{T:data_icn_descr}
\end{table}

\subsection{TLT-school}

In Trentino, an autonomous region in northern Italy, the linguistic competence of Italian students has been assessed in recent years through proficiency tests in both English and German~\cite{gretter2020}, involving about 3000 students ranging from 9 to 16 years old, belonging to four different school grades ($5^{th}$, $8^{th}$, $10^{th}$, $11^{th}$) and three CEFR proficiency levels (A1, A2, B1). 
Since our experiments are conducted only on the B1 section of the English spoken parts of the corpus, we will not describe the section concerning the German section, as their analysis goes beyond the scope of this paper.

After eliminating the answers containing only silence or non-speech background, 
the spoken section is composed of 494 responses to 7 small talk questions about everyday life situations.
It is worth mentioning that some answers are characterized by a number of issues (e.g., presence of words belonging to multiple languages or presence of off-topic answers). 
We decided not to eliminate these answers from the data used in the experiments. As regards the manual transcriptions, we did not eliminate the annotations related to spontaneous speech phenomena, such as hesitations and fragments of words, in order not to lose any possibly existing information about pronunciation and fluency, although we acknowledge that they cannot replace these actual speech phenomena completely. Detailed information about the manual transcriptions and other aspects of the corpus can be found in~\cite{gretter2020}. 

As for the automatic speech recognition (ASR) output text, its WER is 35.9\% on the whole spoken test data, whereas it amounts to 41.13\% for the B1 subset we used in our experiments; acoustic and language models are described in~\cite{gretter2019}. 

The total score ranges from 0 to 12 in the spoken section and consists of the sum of the analytic subscores assigned by human experts for each specific proficiency indicator assigned by the human raters (i.e., relevance, formal correctness, lexical richness and complexity, pronunciation, fluency, and communicative effectiveness). For each indicator, human raters could choose 0, 1 or 2 points. 

For this dataset, we treated proficiency assessment as a regression task when predicting both the holistic score from 0 to 12 and the analytic subscores ranging from 0 to 2.

\begin{figure}[h!]
\includegraphics[scale=0.6]{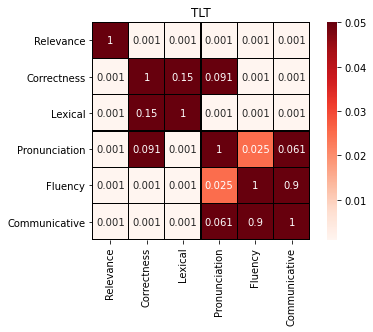}
\caption{Heatmap of the results of the post-hoc Nemenyi test on the TLT-school dataset.}
\label{nemenyi_test}
\centering
\end{figure}

In order to verify that these subscores are, in fact, targeting different aspects of proficiency, we investigated the relationships between them with a repeated measures design. Since our data violate both the sphericity and normality assumptions required for rANOVA, we performed the Friedman test \cite{friedman}, which is the non-parametric equivalent of rANOVA and determines whether there are any statistically significant differences in ranks between the distributions of multiple paired groups. As we obtained a significant \emph{p}-value, we found that there are significant differences among the subscores. In order to determine exactly which pairs of subscores are significantly different, we performed post-hoc multiple comparisons using the Nemenyi test \cite{nemenyi} (see Figure~\ref{nemenyi_test}). All paired comparisons show significant differences (\emph{p}-value\textless{0.05}), except the pairs formal correctness-lexical richness and complexity, formal correctness-pronunciation, pronunciation-communicative effectiveness, and fluency-communicative effectiveness. The absence of significant differences between the subscores related to correctness and those related to lexical richness and complexity seem to be consistent with the fact that: a) a poorer and simpler lexis tends to have fewer errors and b) the human expert may have actually overlapped the two indicators due to the presence of lexical errors, which can be connected either to poor formal correctness or bad use of vocabulary. Similarly, pronunciation and correctness do not show significant differences, because pronunciation errors might have been incorporated in one or the other indicator. On the other hand, the indicator related to communicative effectiveness intersects almost by definition with those related to fluency and pronunciation. This has been particularly true in recent years, as pronunciation tends to be assessed in terms of general goals such as intelligibility~\cite{levis2018} and communicative effectiveness~\cite{pennington2019} rather than closeness to native English, especially in English as an International Language (EIL) contexts~\cite{low2014}.
Regardless of all these considerations, one also has to take into account that halo effect~\cite{myford2003} might bias the scores to a certain extent.

Since every utterance was scored by only one expert, it was not possible to evaluate any kind of agreement among experts. Note that the CEFR levels were assigned before the tests and should be considered as expected proficiency levels, whereas the test scores are effectively representing each learner's performance in the exam. We divided the data into three sets: a training set of 322, a development set of 85 and a test set of 87 samples.

\begin{table}[h!]

\begin{center}
\begin{tabular}{ l | c | c | c | c }

   & \textbf{Train} & \textbf{Dev} & \textbf{Test} & \textbf{Total} \\ 

\hline
\hline
0-3 & 74   & 14 & 27 & 115 \\ \hline
3-6 & 73  & 20 & 17 & 110 \\ \hline
6-9 & 77   & 20 & 11 & 108 \\ \hline 
9-12 & 98 & 31 & 32 & 161 \\ \hline \hline
Total & 322 & 85 & 87 & 494 \\ \hline

\end{tabular}
\end{center}
\caption{Number of answers for each score range in TLT-school.}
\label{T:tlt_descr}
\end{table}

\section{Model architectures}
\label{sec:pagestyle}
\subsection{wav2vec2-based graders}

Wav2vec 2.0 encodes speech audio through a multilayer convolutional neural network (CNN). After encoding, masking is applied to spans of the resulting latent representations, which are fed into a transformer to build contextualised representations. Gumbel softmax is employed to compute the contrastive loss on which the model is trained, and speech representations are learned from this training. For our models, we initialised the configuration and processor from a version provided by the HuggingFace Transformer Library~\cite{huggingface}\footnote{\url{huggingface.co/patrickvonplaten/wav2vec2-base}}.
After the learners' answers are fed into the model, wav2vec 2.0 provides contextualised representations. In order to handle representations of different audio lengths, we used a mean pooling method to concatenate 3D representations into 2D representations. These are subsequently passed through a Dense layer of 768 units, a Dropout layer and, finally, through an output layer. We tried different architectures and hyperparameters and, finally, we opted for those described in the following paragraphs.

\textbf{ICNALE:} the task is multi-class classification, therefore the output layer has 5 units and softmax as activation function. The grader uses cross entropy as loss function. Training uses the AdamW optimiser \cite{adamw} with batch size 4, gradient accumulation step 2, dropout 0.2, and learning rate 1e-5. The grader is trained for 8 epochs.

\textbf{TLT-school - holistic score:} in this case, assessment is treated as a regression task, therefore the output layer has 1 unit and a linear activation function. The loss function is mean squared error (MSE). The grader is trained for 12 epochs using AdamW optimiser with batch size 4, gradient accumulation step 2, dropout 0.2, and learning rate 5e-5.

\textbf{TLT-school - analytic subscores:} we trained 6 different graders for each aspect of proficiency. The batch size and gradient accumulation steps are the same as the holistic grader, whereas the other hyperparameters are reported in Table \ref{T:specific_graders}.

\begin{table}[h!]

\begin{center}
\begin{tabular}{ l | c | c | c }

   & \textbf{Epochs} & \textbf{Learning rate} & \textbf{Dropout} \\ 
\hline

\hline
Relevance & 13   & 5e-6 & 0.1  \\ \hline
Correctness & 19  & 2e-6 & 0.1  \\ \hline
Lexical & 12   & 4e-6 & 0.1 \\ \hline 
Pronunciation & 8 & 1e-5 & - \\ \hline
Fluency & 6 & 8e-6 & 0.1 \\ \hline
Communicative & 10 & 1e-5 & 0.1 \\ \hline

\end{tabular}
\end{center}
\caption{Hyperparameters of the individual wav2vec2-based graders.}
\label{T:specific_graders}
\end{table}

As has been said, the first component of wav2vec 2.0 consists of a stack of CNN layers that are used to extract acoustically meaningful - but contextually independent - features from the raw speech signal. This part of the model has already been sufficiently trained during pretraining and does not need to be fine-tuned. Therefore, we froze the parameters of the feature extractor for all our experiments.

\subsection{BERT-based graders}

The baseline systems used for comparison are BERT-based models in the version provided by the HuggingFace Transformer Library~\cite{huggingface} \footnote{\url{huggingface.co/bert-base-uncased}}. The models take a sequence of token embeddings, i.e., of the answers provided by the learners as inputs. Each token is transformed into a vector representation and then passed to BERT's encoder layer. We use the [CLS] token state and feed it to a classification or regression head, depending on the nature of the task. Similarly to what we did with our wav2vec2-based models, we kept the BERT layer frozen. We tried various hyperparameters and architectures and we chose the ones described in the following paragraphs.

\textbf{ICNALE:} the classification head consists of a stack of three Dense layers of 768 units, a stack of three Dense layers of 128 units, and an output layer of 5 units with softmax as activation function. The model is trained for 600 epochs with batch size set to 256 using Adam optimiser~\cite{kingma2014} with learning rate set to 5e-5 and cross entropy as loss function. The maximum sequence length is 256.

\textbf{TLT-school - holistic score:} the regression head has the same intermediate layers as the model used on ICNALE and an output layer of 1 unit with linear activation function. The model is trained for 800 epochs on the manual transcriptions and for 150 epochs on the ASR transcriptions. The batch size is set to 256 and the maximum sequence length to 64. Training uses Adam optimiser with learning rate 2e-5 and dropout 0.2. 

\textbf{TLT-school - analytic subscores:} analogously to what we did with the wav2vec2-based graders, we trained a BERT-based grader for each proficiency indicator. The architecture and hyperparameters of the individual graders are the same as the holistic grader except for the dropout rate of the fluency grader, which is set to 0.4.

The performance of the graders trained on ICNALE is evaluated using accuracy and weighted F1 score, whereas for the evaluation of both the holistic and individual graders trained on TLT-school we considered Pearson's correlation coefficient (PCC), Spearman's rank coefficient (SRC) and mean squared error (MSE).

\section{Experiments and results}
\label{sec:typestyle}

\subsection{Results on ICNALE}

\begin{figure}
\centering

\subfloat{\includegraphics[width=6cm]{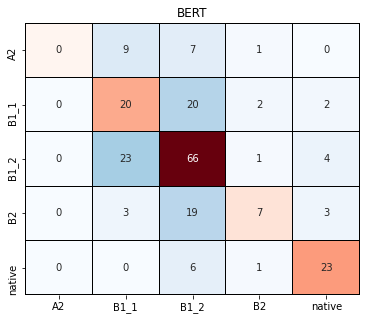}\label{fig:BERT_conf}}

\subfloat{\includegraphics[width=6cm]{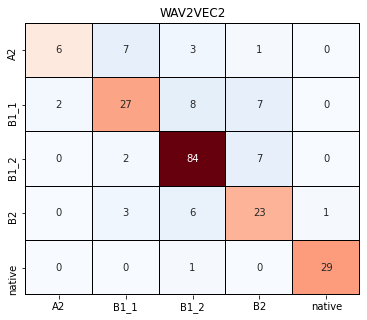}\label{fig:wav_conf}}

\caption{Confusion matrices of CEFR proficiency levels for the two graders (predicted labels on X-axis, true labels on the Y-axis) on the ICNALE test set.}
\label{fig:confusion_matrs}
\end{figure}

We started our series of experiments from the multi-class classifiers trained on ICNALE. First, we trained and evaluated the BERT-based baseline grading system. Secondly, we tried the wav2vec2-based grader. Table \ref{T:results_ICNALE} shows the results of the two graders in terms of accuracy and weighted F1 score on the ICNALE test set. Figure~\ref{fig:confusion_matrs} illustrates the confusion matrices of each CEFR proficiency level for the two graders. As can be seen, the wav2vec-based grader significantly outperforms the BERT baseline across all proficiency levels. Specifically, it performs best on B1\_2 and on the class related to native speakers. While the reason for its results on the latter can be attributed to a clear gap between L1 and low/mid levels of L2 English, such as the ones featured in the dataset, the performance on the first is probably due to the greater amount of training data available compared to the other CEFR proficiency levels. Similarly but conversely, this can be inferred from the relatively worse performance on A2, which is the least represented proficiency level (see Table \ref{T:data_icn_descr}).

\begin{table}[h!]

\begin{center}
\begin{tabular}{ l | c | c }

   & \textbf{Accuracy(\%)} & \textbf{Weighted F1}\\ 
\hline

\hline
BERT & 53.45 & 0.50 \\ \hline
wav2vec2 & 77.88 & 0.77 \\ \hline

\end{tabular}
\end{center}
\caption{Results on the ICNALE test set of the BERT-based and wav2vec2-based graders in terms of accuracy and weighted F1 score.}
\label{T:results_ICNALE}
\end{table}

\subsection{Results on TLT-school}

We continued our experiments using the TLT-school data, starting from the holistic graders. In this case, we compared the performance of the wav2vec2-based model and the BERT-based model using the manual transcriptions and the ASR output text. Table \ref{T:results_TLT_holistic} reports the results on the TLT-school test set. It can be observed that the wav2vec2-based approach significantly outperforms BERT both on the manual and ASR transcriptions across all metrics.

\begin{table}[h!]

\begin{center}
\begin{tabular}{ l | c | c | c}

   & \textbf{PCC} & \textbf{SRC} & \textbf{MSE}\\ 
\hline

\hline
BERT-ASR & 0.749 & 0.743 & 9.877 \\ \hline
BERT-manual & 0.857  & 0.863 & 6.110  \\ \hline
wav2vec2 & 0.927  & 0.933 & 2.297  \\ \hline

\end{tabular}
\end{center}
\caption{Results on TLT-school test set (holistic score) of the BERT-based grader (manual and ASR transcriptions) and the wav2vec2-based grader in terms of PCC, SRC and MSE.}
\label{T:results_TLT_holistic}
\end{table}

\begin{figure}
\centering
\subfloat{\includegraphics[width=8cm]{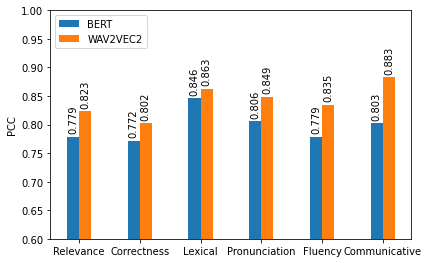}\label{fig:multi_PCC}}

\subfloat{\includegraphics[width=8cm]{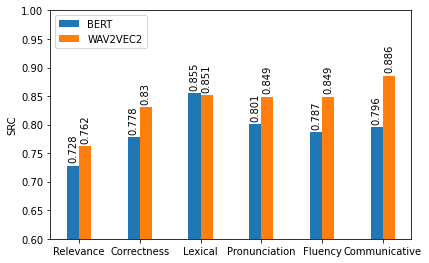}\label{fig:multi_SRC}}

\caption[]{Comparison of the BERT-based and wav2vec2-based individual graders in terms of PCC and SRC.}
\label{fig:comparison_specific}
\end{figure}

Subsequently, in order to verify the impact of our two approaches on individual aspects of proficiency, we trained our models on the analytic subscores of each proficiency indicator in the TLT-school test set: relevance, formal correctness, lexical richness and complexity, pronunciation, fluency, and communicative effectiveness. For this part of our experiments, we did not consider the ASR transcriptions, but we only focused on the upper bound provided by the manual transcriptions. We report the results in Table \ref{fig:comparison_specific}. As can be seen, the use of wav2vec 2.0 significantly improves on the performance of BERT.
In particular, as expected, the more exquisitely speech-related analytic subscores (i.e., pronunciation, fluency, and communicative effectiveness) are best predicted using wav2vec 2.0, despite the presence of proxies of fluency and pronunciation in the manual transcriptions. Furthermore, the wav2vec2-based graders remarkably outperform the BERT baselines when predicting the formal correctness subscore and the relevance subscore. The only subscore on which the predictions of both approaches appear fairly aligned is the one related to lexical richness and complexity, despite the BERT baseline performance being lower in terms of PCC and MSE. This is quite an expectable result, given that BERT is trained on a large quantity of textual data, and lexical richness and complexity are competences that should be typically constructed and evaluated starting from text, be it written or spoken.

\begin{figure}[h!]
\includegraphics[scale=0.6]{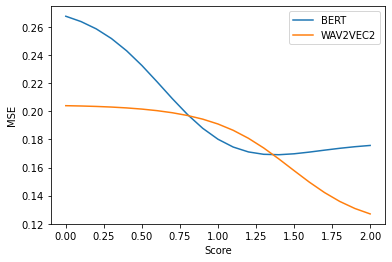}
\caption{MSE variation of the wav2vec2-based and BERT-based (manual transcription) graders across scores for lexical richness and complexity.}
\label{mse_var_lex}
\centering
\end{figure}

Given such results, we continued our analysis focusing on this indicator. Specifically, we wanted to analyse the performance of the wav2vec2- and BERT-based graders across scores, in order to understand whether these two approaches perform differently across proficiency and therefore could be complementary to each other. Figure~\ref{mse_var_lex} illustrates the MSE variation across scores applying Gaussian kernel smoothing with sigma set to 0.5.

We found that the BERT-based model performs moderately better in the middle scores, whilst the approach including wav2vec 2.0 has a significantly lower MSE for the lowest and highest scores. To verify their complementarity, we merged them using: 
a) a {\em shallow} combination: 
we calculated the average of the scores predicted by each grader; 
b) a {\em deep} combination: 
we concatenated the two hidden representations in BERT and wav2vec 2.0,
and we fed them through a small network consisting of a Dense layer of 16 units, Dropout layer with dropout rate set to 0.5 and a final output layer. The resulting network was trained for 3000 epochs with learning rate 5e-5 and batch size 512. In both cases, an interesting improvement across all considered metrics is observed, as shown in Table~\ref{T:comparison_lex}.

\begin{table}[h!]

\begin{center}
\begin{tabular}{ l | c | c | c}

   & \textbf{PCC} & \textbf{SRC} & \textbf{MSE}\\ 
\hline

\hline
BERT & 0.846 & 0.855 & 0.217 \\ \hline
wav2vec2 & 0.863  & 0.851 & 0.178  \\ \hline
BERT+wav2vec2 (shallow) & 0.885   & 0.876  & 0.164   \\ \hline
BERT+wav2vec2 (deep) & 0.883 & 0.864 & 0.166 \\
\hline

\end{tabular}
\end{center}
\caption{Results on TLT-school test set (lexical richness and complexity) of the BERT-based (manual transcriptions), the wav2vec2-based graders, and their combinations in terms of PCC, SRC and MSE.}
\label{T:comparison_lex}
\end{table}

\section{Conclusions and future works}
\label{sec:majhead}

Transcriptions of L2 speaking tests are not always easy to obtain, and, even when they are available through ASR systems, they generally contain errors and do not provide information about strictly speech-related aspects of proficiency, such as pronunciation and fluency. This paper considers whether it is possible to use wav2vec 2.0 representations to assess L2 spoken English proficiency both holistically and analytically, even when a small quantity of data is available. First, we found that this approach significantly outperforms the BERT baseline system trained on manual transcriptions of the ICNALE dataset in the task of CEFR level classification. Secondly, we investigated the use of wav2vec 2.0 for a regression task on the B1 section of TLT-school targeting holistic scores. In this case, we also obtained significant improvements on the BERT baseline trained on ASR and manual transcriptions. Finally, we tested this approach on subscores related to individual aspects of proficiency (i.e., relevance, formal correctness, lexical richness and complexity, pronunciation, fluency, and communicative effectiveness) using manual transcriptions only, and we found that the wav2vec2-based graders remarkably outperform the BERT-based baseline systems across all proficiency indicators. For lexical richness and complexity, i.e., the only subscore on which the two strategies had a similar performance, we found that two types of combination of the two models show an interesting improvement, suggesting a certain complementarity. Future works will involve other approaches for comparison, e.g., a grader with hand-crafted features, and other types of combination, considering both shallow and deep fusion methods. Furthermore, specific types of combinations will be investigated according to each aspect of proficiency, e.g., a concatenation of the question prompt and learner's answer for the subscore related to relevance along the lines of~\cite{qian2018}. In this work, we have focused on answer-level assessment, but we plan to investigate whether it is possible to evolve this approach in order to provide feedback about specific parts of an answer, e.g., analysing the local attention representations. Finally, we did not fine-tune wav2vec 2.0 on L2 English data before using it for proficiency assessment, therefore we envisage the investigation of this additional step to further improve our results.


\bibliographystyle{IEEEbib}
\bibliography{refs}

\begin{thebibliography}{10}

\bibitem{zhang2013}
M.~Zhang,
\newblock ``Contrasting automated and human scoring of essays,''
\newblock {\em R\&D Connections}, , no. 21, pp. 1--11, 2013.

\bibitem{iwashita2008}
N.~Iwashita, A.~Brown, T.~McNamara, and S.~O'Hagan,
\newblock ``Assessed levels of second language speaking proficiency: How
  distinct?,''
\newblock {\em Applied Linguistics}, vol. 29, no. 1, pp. 24--49, 2008.

\bibitem{dejong2012}
{N.H. De Jong}, {M.P. Steinel}, {A.F. Florijn}, R.~Schoonen, and {J.H.
  Hulstijn},
\newblock ``Facets of speaking proficiency,''
\newblock {\em Studies in Second Language Acquisition}, , no. 34, pp. 5--34,
  2012.

\bibitem{strik1999automatic}
H.~Strik and C.~Cucchiarini,
\newblock ``Automatic assessment of second language learners' fluency,''
\newblock in {\em Proc. International Congress of Phonetic Sciences (ICPhS)
  1999}, 1999.

\bibitem{chen}
L.~Chen, K.~Evanini, and X.~Sun,
\newblock ``Assessment of non-native speech using vowel space
  characteristics,''
\newblock in {\em Proc. 2010 IEEE Spoken Language Technology Workshop}, 2010,
  pp. 139--144.

\bibitem{coutinho2016assessing}
E.~Coutinho, F.~H{\"o}nig, Y.~Zhang, S.~Hantke, A.~Batliner, E.~N{\"o}th, and
  B.~Schuller,
\newblock ``Assessing the prosody of non-native speakers of {English}: Measures
  and feature sets,''
\newblock in {\em Proc. 10th International Conference on Language Resources and
  Evaluation (LREC'16)}, 2016.

\bibitem{bhat2015automatic}
S.~Bhat and S.~Yoon,
\newblock ``Automatic assessment of syntactic complexity for spontaneous speech
  scoring,''
\newblock {\em Speech Communication}, vol. 67, pp. 42--57, 2015.

\bibitem{muller2009automatically}
P.~M{\"u}ller, F.~De~Wet, {C. Van Der Walt}, and T.~Niesler,
\newblock ``Automatically assessing the oral proficiency of proficient {L2}
  speakers.,''
\newblock in {\em Proc. Workshop on Speech and Language Technology for
  Education (SLaTE)}, 2009, pp. 29--32.

\bibitem{crossley2013applications}
S.~Crossley and D.~McNamara,
\newblock ``Applications of text analysis tools for spoken response grading,''
\newblock {\em Language Learning \& Technology}, vol. 17, no. 2, pp. 171--192,
  2013.

\bibitem{wang2018towards}
Y.~Wang, {M.J.F. Gales}, {K.M. Knill}, K.~Kyriakopoulos, A.~Malinin, {R.C. van
  Dalen}, and M.~Rashid,
\newblock ``Towards automatic assessment of spontaneous spoken {English},''
\newblock {\em Speech Communication}, vol. 104, pp. 47--56, 2018.

\bibitem{liu2020dolphin}
Z.~Liu, G.~Xu, T.~Liu, W.~Fu, Y.~Qi, W.~Ding, Y.~Song, C.~Guo, C.~Kong,
  S.~Yang, et~al.,
\newblock ``Dolphin: a spoken language proficiency assessment system for
  elementary education,''
\newblock in {\em Proc. The Web Conference 2020}, 2020, pp. 2641--2647.

\bibitem{chen2018end}
L.~Chen, J.~Tao, S.~Ghaffarzadegan, and Y.~Qian,
\newblock ``End-to-end neural network based automated speech scoring,''
\newblock in {\em Proc. 2018 IEEE International Conference on Acoustics, Speech
  and Signal Processing (ICASSP)}, 2018, pp. 6234--6238.

\bibitem{takai2020deep}
K.~Takai, P.~Heracleous, K.~Yasuda, and A.~Yoneyama,
\newblock ``Deep learning-based automatic pronunciation assessment for second
  language learners,''
\newblock in {\em Proc. International Conference on Human-Computer
  Interaction}, 2020, pp. 338--342.

\bibitem{cheng2020asr}
S.~Cheng, Z.~Liu, L.~Li, Z.~Tang, D.~Wang, and {T.F. Zheng},
\newblock ``{ASR-Free} pronunciation assessment,''
\newblock in {\em Proc. Interspeech 2020}, 2020, pp. 3047--3051.

\bibitem{kyriakopoulos2018deep}
K.~Kyriakopoulos, {K.M. Knill}, and {M.J.F. Gales},
\newblock ``A deep learning approach to assessing non-native pronunciation of
  {English} using phone distances,''
\newblock in {\em Proc. Interspeech 2018}, 2018, pp. 1626--1630.

\bibitem{kyriakopoulos2019deep}
K.~Kyriakopoulos, {K.M. Knill}, and {M.J.F. Gales},
\newblock ``A deep learning approach to automatic characterisation of rhythm in
  non-native {English} speech,''
\newblock in {\em Proc. Interspeech 2019}, 2019, pp. 1836--1840.

\bibitem{wang2021}
X.~Wang, K.~Evanini, Y.~Qian, and M.~Mulholland,
\newblock ``Automated scoring of spontaneous speech from young learners of
  english using transformers,''
\newblock in {\em 2021 IEEE Spoken Language Technology Workshop (SLT)}, 2021,
  pp. 705--712.

\bibitem{raina2020universal}
V.~Raina, {M.J.F. Gales}, and {K.M. Knill},
\newblock ``Universal adversarial attacks on spoken language assessment
  systems,''
\newblock in {\em Proc. Interspeech 2020}, 2020, pp. 3855--3859.

\bibitem{banno2022view}
S.~Bann{\`o}, B.~Balusu, M.~J.~F. Gales, K.~M. Knill, and K.~Kyriakopoulos,
\newblock ``View-specific assessment of {L2} spoken {English},''
\newblock in {\em Proc. Interspeech 2022}, 2022, pp. 4471--4475.

\bibitem{lu2022}
Y.~Lu, S.~Bann{\`o}, and M.J.F. Gales,
\newblock ``On assessing and developing spoken {'}grammatical error
  correction{'} systems,''
\newblock in {\em Proceedings of the 17th Workshop on Innovative Use of NLP for
  Building Educational Applications (BEA 2022)}, 2022, pp. 51--60.

\bibitem{baevski}
A.~Baevski, H.~Zhou, A.~Mohamed, and M.~Auli,
\newblock ``wav2vec 2.0: A framework for self-supervised learning of speech
  representations,''
\newblock in {\em NeurIPS 2020}, 2020, pp. 1--12.

\bibitem{hsu}
{W.N. Hsu}, A.~Sriram, A.~Baevski, T.~Likhomanenko, Q.~Xu, V.~Pratap, J.~Kahn,
  A.~Lee, R.~Collobert, G.~Synnaeve, and M.~Auli,
\newblock ``Robust wav2vec 2.0: Analyzing domain shift in self-supervised
  pre-training,''
\newblock in {\em Proc. Interspeech 2021}, 2021, pp. 721--725.

\bibitem{yang21c}
{S.-W Yang}, {P.-H. Chi}, {Y.-S. Chuang}, {C.-I. J. Lai}, K.~Lakhotia, {Y.Y.
  Lin}, {A.T. Liu}, J.~Shi, X.~Chang, {G.-T. Lin}, {T.-H. Huang}, {W.-C.
  Tseng}, {K.-T. Lee}, {D.-R. Liu}, Z.~Huang, S.~Dong, {S.-W. Li}, {S.
  Watanabe}, {A. Mohamed}, and {H.-Y. Lee},
\newblock ``{SUPERB: Speech Processing Universal PERformance Benchmark},''
\newblock in {\em Proc. Interspeech 2021}, 2021, pp. 1194--1198.

\bibitem{shah}
J.~{Shah}, Y.K. {Singla}, C.~{Chen}, and R.~{Ratn Shah},
\newblock ``{What all do audio transformer models hear? Probing Acoustic
  Representations for Language Delivery and its Structure},''
\newblock {\em arXiv e-prints}, p. arXiv:2101.00387, 2021.

\bibitem{peng21}
L.~Peng, K.~Fu, B.~Lin, D.~Ke, and J.~Zhan,
\newblock ``{A Study on Fine-Tuning wav2vec2.0 Model for the Task of
  Mispronunciation Detection and Diagnosis},''
\newblock in {\em Proc. Interspeech 2021}, 2021, pp. 4448--4452.

\bibitem{wu21}
M.~Wu, K.~Li, {W.-K. Leung}, and H.~Meng,
\newblock ``{Transformer Based End-to-End Mispronunciation Detection and
  Diagnosis},''
\newblock in {\em Proc. Interspeech 2021}, 2021, pp. 3954--3958.

\bibitem{xu21}
X.~Xu, Y.~Kang, S.~Cao, B.~Lin, and L.~Ma,
\newblock ``{Explore wav2vec 2.0 for Mispronunciation Detection},''
\newblock in {\em Proc. Interspeech 2021}, 2021, pp. 4428--4432.

\bibitem{eesung}
E.~Kim, J.-J. Jeon, H.~Seo, and H.~Kim,
\newblock ``Automatic pronunciation assessment using self-supervised speech
  representation learning,''
\newblock in {\em Proc. Interspeech 2022}, 2022, pp. 1411--1415.

\bibitem{devlin2018}
J.~Devlin, M.~Chang, L.~Kenton, and K.~Toutanova,
\newblock ``{BERT: Pre-training of Deep Bidirectional Transformers for Language
  Understanding},''
\newblock {\em arXiv e-prints}, p. arXiv:1810.04805, 2018.

\bibitem{ishikawa2011}
S.~Ishikawa,
\newblock ``A new horizon in learner corpus studies: The aim of the {ICNALE}
  project,''
\newblock in {\em Corpora and language technologies in teaching, learning and
  research}, G.~Weir, S.~Ishikawa, and K.~Poonpon, Eds., pp. 3--11. University
  of Strathclyde Press, 2011.

\bibitem{cefr2001}
{Council of Europe},
\newblock {\em Common European Framework of Reference for Languages: Learning,
  Teaching, Assessment},
\newblock Cambridge University Press, Cambridge, 2001.

\bibitem{zhou19}
Z.~Zhou, S.~Vajjala, and {S.V. Mirnezami},
\newblock ``{Experiments on Non-native Speech Assessment and its
  Consistency},''
\newblock in {\em NLP4CALL 2019}, 2019, pp. 86--92.

\bibitem{banno2022}
S.~Bann{\`o} and M.~Matassoni,
\newblock ``Cross-corpora experiments of automatic proficiency assessment and
  error detection for spoken {E}nglish,''
\newblock in {\em Proceedings of the 17th Workshop on Innovative Use of NLP for
  Building Educational Applications (BEA 2022)}, 2022, pp. 82--91.

\bibitem{gretter2020}
R.~Gretter, M.~Matassoni, S.~Bann{\`o}, and D.~Falavigna,
\newblock ``{TLT}-school: a corpus of non native children speech,''
\newblock in {\em Proceedings of the 12th Language Resources and Evaluation
  Conference}, 2020.

\bibitem{gretter2019}
R.~Gretter, M.~Matassoni, K.~Allgaier, S.~Tchistiakova, and D.~Falavigna,
\newblock ``Automatic assessment of spoken language proficiency of non-native
  children,''
\newblock in {\em IEEE International Conference on Acoustics, Speech and Signal
  Processing}, 2019.

\bibitem{friedman}
M.~Friedman,
\newblock ``The use of ranks to avoid the assumption of normality implicit in
  the analysis of variance,''
\newblock {\em Journal of the American Statistical Association}, vol. 32, no.
  200, pp. 675--701, 1937.

\bibitem{nemenyi}
{P.B. Nemenyi},
\newblock {\em Distribution-free multiple comparisons},
\newblock Ph.D. thesis, Princeton University, 1963.

\bibitem{levis2018}
J.M. Levis,
\newblock {\em Intelligibility, Oral Communication, and the Teaching of
  Pronunciation},
\newblock Cambridge University Press, Cambridge, 2018.

\bibitem{pennington2019}
{M.C. Pennington} and P.~Rogerson-Revell,
\newblock {\em English Pronunciation Teaching and Research: Contemporary
  Perspectives},
\newblock Palgrave Macmillan, London, 2019.

\bibitem{low2014}
E.-L. Low,
\newblock {\em Pronunciation for English as an International Language: from
  Research to Practice},
\newblock Routledge, London, 2014.

\bibitem{myford2003}
{C.M. Myford and E.W. Wolfe},
\newblock ``Detecting and measuring rater effects using many-facet rasch
  measurement: Part i,''
\newblock {\em Journal of Applied Measurement}, vol. 4, no. 4, pp. 386--422,
  2003.

\bibitem{huggingface}
T.~Wolf, L.~Debut, V.~Sanh, J.~Chaumond, C.~Delangue, A.~Moi, P.~Cistac,
  T.~Rault, R.~Louf, M.~Funtowicz, J.~Davison, S.~Shleifer, P.~von Platen,
  C.~Ma, Y.~Jernite, J.~Plu, C.~Xu, T.~Le~Scao, S.~Gugger, M.~Drame, Q.~Lhoest,
  and A.~Rush,
\newblock ``Transformers: State-of-the-art natural language processing,''
\newblock in {\em Proceedings of the 2020 Conference on Empirical Methods in
  Natural Language Processing: System Demonstrations}, 2020, pp. 38--45.

\bibitem{adamw}
I.~Loshchilov and F.~Hutter,
\newblock ``Decoupled weight decay regularization,''
\newblock in {\em ICLR 2017}, 2017.

\bibitem{kingma2014}
D.P. Kingma and J.~Ba,
\newblock ``Adam: a method for stochastic optimization,''
\newblock in {\em International Conference on Learning Representations}, 2014.

\bibitem{qian2018}
Y.~Qian, R.~Ubale, M.~Mulholland, K.~Evanini, and X.~Wang,
\newblock ``A prompt-aware neural network approach to content-based scoring of
  non-native spontaneous speech,''
\newblock in {\em 2018 IEEE Spoken Language Technology Workshop (SLT)}, 2018,
  pp. 979--986.

\end{thebibliography}

\end{document}